\documentclass[preprint,review, 12pt]{elsarticle}

\usepackage{epsfig}
\usepackage{amssymb}
\usepackage{amsthm}
\usepackage{amsmath}
\usepackage{url,hyperref}
\usepackage{lineno}
\usepackage{enumitem}
\usepackage{setspace}

\usepackage{algorithm,algorithmic}

\journal{IPL}

\begin{document}
\begin{frontmatter}

\title{An incremental linear-time learning algorithm for the Optimum-Path Forest classifier}

\author{Moacir Ponti}
\author{Mateus Riva}

\address{Instituto de Ci\^encias Matem\'aticas e de Computa\c{c}\~ao, Universidade de S\~ao Paulo -- S\~ao Carlos, SP 13566-590 Brazil}

\begin{abstract}
We present a classification method with incremental capabilities based on the Optimum-Path Forest classifier (OPF). The OPF considers instances as nodes of a fully-connected training graph, arc weights represent distances between two feature vectors. Our algorithm includes new instances in an OPF in linear-time, while keeping similar accuracies when compared with the original quadratic-time model.
\end{abstract}

\end{frontmatter}


\section{Introduction}
\label{s.introduction}
The optimum-path forest (OPF) classifier~\cite{Papa2012} a classification method that can be used to build simple, multiclass and parameter independent classifiers. One possible drawback of using the OPF classifier in learning scenarios in which there is need to constantly update the model, is its quadratic training time. Let a training set be composed of $n$ examples, the OPF training algorithm runs in $O(n^2)$. Some efforts were made to mitigate such running time by using several OPF classifiers trained with ensembles of reduced training sets~\cite{PontiRossi2013} and fusion using split sets using multi-threading~\cite{Ponti2011a}. Also, recent work developed strategies to speed-up the training algorithm by taking advantage of data structures such as~\cite{Papa2012,Iwashita2014}. However, an OPF-based method with incremental capabilities is still to be investigated, since sub-quadratic algorithms are important in many scenarios~\cite{Berenbrink2014}. 

Incremental learning is a machine learning paradigm in which the classifier changes and adapts itself to include new examples that emerged after the initial construction of the classifier~\cite{Geng2015}. As such, an incremental-capable classifier has to start with an incomplete \textit{a priori} dataset and include successive new data without the need to rebuild itself. In~\cite{Iwashita2014} the authors propose an alternative OPF algorithm which is more efficient to retrain the model, but their algorithm is not incremental. Also, the empirical evidence shows that the running time is still quadratic, although with a significantly smaller constant. In this paper we describe an algorithm that can include new examples individually (or in small batches) in an already build model, which is a different objective when compared to~\cite{Iwashita2014} and \cite{Papa2012}. In fact, we already used the improvements proposed by~\cite{Papa2012}. Therefore our new algorithm {\bf does not compete}, but rather can be used as a complement for those variants. 

Because OPF is based on the Image Foresting Transform for which there is a differential algorithm available (DIFT)~\cite{Falcao2004b}, it would be a natural algorithm to try. However, DIFT is an image processing algorithm and includes all new pixels/nodes as prototypes, which would progressively convert the model into a 1-Nearest Neighbour classifier. Therefore we propose an alternative solution that maintains the connectivity properties of optimum-path trees.

Our OPF-Incremental (OPFI) is inspired in graph theory methods to update minimum spanning trees~\cite{ChinHouck1978} and minimal length paths~\cite{Ausiello1991} in order to maintain the graph structure and thus the learning model. We assume there is an initial model trained with the original OPF training, and then perform several inclusions of new examples appearing over time. This is an important feature since models should be updated in an efficient way in order to comply with realistic scenarios. Our method will be useful everywhere the original OPF is useful, along with fulfilling incremental learning requirements.

\section{OPF Incremental (OPFI)}
\label{s.opfincremental}

The optimum-path forest (OPF) classifier~\cite{Papa2012} interprets the instances (examples) as the nodes (vertices) of a graph. The edges connecting the vertices are defined by some adjacency relation between the examples, weighted by a distance function. It is expected that training examples from a given class will be connected by a path of nearby examples. Therefore the model that is learned by the algorithm is composed by several trees, each tree is a minimum spanning tree (MST) and the root of each tree is called prototype. 

Our OPF incremental updates an initial model obtained by the original OPF training by using the minimum-spanning tree properties in the existing optimum-path forest. Provided this initial model, our algorithm is able to include a new instance in linear-time. Note that in incremental learning scenarios it is typical to start with an incomplete training set, often presenting a poor accuracy due to the lack of a sufficient sample. 

Our solution works by first classifying the new example using the current model. Because the label of the classified example is known, it is possible to infer if it has been conquered by a tree of the same class (i.e. it was correctly classified) or not. We also know which node was responsible for the conquest, i.e. its {\bf predecessor}. Using this knowledge, we the have three possible cases:
\begin{enumerate}
\item \textbf{Predecessor belongs to the same class and is not a prototype}: the new example is inserted in the predecessor's tree, maintaining the properties of a minimum spanning tree.
\item \textbf{Predecessor belongs to the same class and is a prototype}: we must discover if the new example will take over as prototype. If so, the new prototype must reconquer the tree; otherwise, it is inserted in the tree as in the first case.
\item \textbf{Predecessor belongs to another class}: the new example and its predecessor become prototypes of a new tree. The new example will be root of an new tree; while the predecessor will begin a reconquest of its own tree, splitting it in two.
\end{enumerate}

\begin{figure}[htpb]
\begin{center}
\begin{tabular}{ccc} 
      \hline \\ [-1.0ex]
      \includegraphics[width=0.25\linewidth]{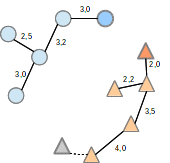} &
      \includegraphics[width=0.25\linewidth]{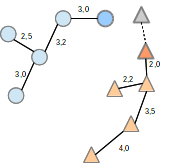} &
      \includegraphics[width=0.31\linewidth]{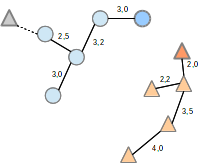}\\
      (a)  & (b) & (c) \\[2pt]
      \hline
\end{tabular}
\caption{OPF-Incremental cases when adding a new example (a gray triangle): (a) conquered by a tree of the same class through a non-prototype, (b) conquered by a tree of the same class through a prototype, (c) conquered by a tree of a distinct class.}
\label{fig:OPFI}
\end{center}
\end{figure}

The Figure~\ref{fig:OPFI} illustrates the three cases when an element of the 'triangle' class is inserted in the OPF.

The classification and insertion of new elements is described on Algorithm~\ref{alg:class_opfi} and shows the high-level solution described above.

\begin{algorithm}
{\small
	\caption{OPF-Incremental insertion}
	\label{alg:class_opfi}
	\begin{algorithmic}[1]	
	\REQUIRE a previously trained OPF model $T$ with $n$ vertices; new instances to be included $Z[1...b]$.
	
	    \STATE OPF\_Classify(Z, T)  // as in \cite{Papa2012}
	    
	    \FOR {$i \gets 1$ to $b$ \text{ (each new example)} }
	    
	        \IF{$Z[i].label = Z[i].truelabel$}
	            \IF{$Z[i].pred$ is $prototype$}
	                \STATE recheckPrototype($Z[i]$,$Z[i].pred$,$T$) \text{ // Algorithm 3 }
	            \ELSE
	                \STATE insertIntoMST($Z[i]$,$Z[i].pred$,$T$) \text{ // Algorithm 2 }
	            \ENDIF
	        \ELSE
	            \STATE $Z[i]$ becomes a prototype
	            \STATE $Z[i].pred$ becomes a prototype
	            \STATE reconquest($Z[i].pred$ $Z[i].pred$, $T$) \text{ // Algorithm 4 }
	        \ENDIF
	    \ENDFOR
	    
		\RETURN $T$
	\end{algorithmic}
}
\end{algorithm}

\begin{algorithm}
{\small
	\caption{OPF-Incremental MST insertion}
	\label{alg:insert_opfi}
	\begin{algorithmic}[1]	
	\REQUIRE $T$ is the graph; $z$ is the new example; $r$ is any vertex in the tree; $t$ is a global variable and is the largest edge in the path between $w$ to $z$, whereas $m$ is the largest edge between $r$ and $z$.
	
	    \STATE mark $r$ "old"
	    \STATE $m \gets (r,z)$
	    \FOR {each vertex $w$ adjacent to $r$}
	        \IF{$w$ is marked "new"}
	            \STATE insertIntoMST($w$, $z$, $T$) \text{ // recursive call }
	            \STATE $k \gets$ the larger of the edges $t$ and $(w,r)$
	            \STATE $h \gets$ the smaller of the edges $t$ and $(w,r)$
	            \STATE $T$ gets the edge $h$
	            \IF{cost of $k <$ cost of $m$}
	                \STATE $m \gets k$
	            \ENDIF
	        \ENDIF
	    \ENDFOR
	    \STATE $t \gets m$
	
		\RETURN $T$
	\end{algorithmic}
}
\end{algorithm}

The minimum spanning tree insertion function, described on Algorithm~\ref{alg:insert_opfi} is an adapted version of the minimum spanning tree updating algorithm proposed by~\cite{ChinHouck1978}. The function for rechecking a prototype, described on Algorithm~\ref{alg:recheck_opfi} takes the distance between the prototype and its pair (the corresponding prototype in the other tree, which edge was cut during the initial phase of classification), and between the new example and said pair. If the new example is closer to the pair, it takes over as prototype and reconquers the tree. Otherwise, it is inserted in the tree. The reconquest function was defined by~\cite{Papa2012}, and described also here on Algorithm~\ref{alg:reconquest_opfi} for clarity.

\begin{algorithm}
{\small
	\caption{OPF-Incremental recheck prototype}
	\label{alg:recheck_opfi}
	\begin{algorithmic}[1]	
	\REQUIRE an input node $Z$ and its predecessor $pred$; a previously trained OPF model $T$; some distance function $dist(\cdot)$.
	
	    \IF{$dist(Z, pred.pair) < dist(pred, pred.pair)$}
	        \STATE $Z$ becomes a prototype
	        \STATE reconquest($Z$, $Z$, $T$) \text{ // Algorithm 4 }
	    \ELSE
	        \STATE insertIntoMST($Z$,$pred$,$T$) \text{ // Algorithm 2 }
	    \ENDIF
	
		\RETURN $T$
	\end{algorithmic}
}
\end{algorithm}

\begin{algorithm}
{\small
	\caption{OPF-Incremental reconquest}
	\label{alg:reconquest_opfi}
	\begin{algorithmic}[1]	
	\REQUIRE an root node $Z$ and its predecessor $pred$; a previously trained OPF model $T$; some distance function $dist(\cdot)$.
	
	    \STATE $Z$ is marked "old"
	
	    \COMMENT{In the first call the root is its own predecessor}
	    
	    \STATE $newPathval \gets dist(Z, pred)$
	    \IF{$newPathval < Z.pathval$}
	        \STATE $Z.pred \gets pred$
	        \STATE $Z.pathval \gets newPathval$
	        \FOR{each adjacent $w$ of $Z$}
	            \IF{$w$ is not "old"}
	                \STATE reconquest($w$,$Z$,$T$) \text{ // recursive call }
	            \ENDIF
	        \ENDFOR
	    \ENDIF
	
		\RETURN $T$
	\end{algorithmic}
}
\end{algorithm}

After the insertion is performed, an ordered list of nodes is updated as in~\cite{Papa2012}. The new example is inserted in its proper position in linear time, thus allowing for the optimisation of the classification step.

When a new instance is inserted, we ensure that its classified label is equal to its true label, which is not always the case as in the original OPF algorithm because a given node can be conquered by a prototype with label different from the true label. Our method differs from the original OPF in this point, but we believe it is important to ensure the label of the new instance because the model is updated upon it and the label plays an important role for instance when a new class appears. Therefore, although our algorithm does not produces a model that is equal to the original OPF, it increments the model by maintaining the optimum path trees properties, rechecking prototypes and including new trees. Those are shown to be enough to achieve classification results that are similar to the original OPF.

\section{Complexity Analysis}
\label{s.analysis}
The complexity of inserting a new example into a model containing a total of $n$ nodes is $\mathcal{O}(n)$, as we demonstrate for each case below.

\paragraph{(i) Predecessor is of another class} splitting a tree is $\mathcal{O}(1)$ since it only requires a given edge to be removed. The reconquest is $\mathcal{O}(n)$, since it goes through each node at most once as described by~\cite{Papa2012};

\paragraph{(ii) Predecessor is of same class and is a prototype} again, the complexity of the reconquest is $\mathcal{O}(n)$. Otherwise, it is an insertion, with complexity $\mathcal{O}(n)$ as described in the case (iii) below.

\paragraph{(iii) Predecessor is of same class and is not a prototype} the complexity of the operation is related to the inclusion of a new example on an existing tree. The complexity is $\mathcal{O}(n)$, or linear in terms of the number of examples, as proof below for Algorithm~\ref{alg:insert_opfi}, showing that the function insertIntoMST() is able to update the MST in linear time. 

\begin{proof}
Let $z$ be the new example conquered by a vertex $r$ on some tree. After executing line 5 of Algorithm~\ref{alg:insert_opfi}, $m$ and $t$ are the largest edges in the paths from $r$ to $z$, and from $w$ (first vertex adjacent to $r$) to $z$ respectively. Because the vertices are numbered in the order that they complete their calls to Algorithm~\ref{alg:insert_opfi} the linearity can be proven by induction.

\paragraph{Base step} The first node to complete its call, say $w'$, must be a leaf node of the graph. Thus, lines 5 to 10 are skipped and $t$ is assigned as $(w', z)$ which is the only edge joining $w'$ and $z$. If $r'$ is a vertex incident to $w'$, it is easy to see that $m = (r', z)$ both before and after the call insertIntoMST($w'$).

\paragraph{Induction steps} When executing line 5, i.e. insertIntoMST($w$), if $w$ is a leaf, again lines 5 and 9 are skipped and $t = (w, z)$. Otherwise, let $x$ be the vertex which is incident to $w$ and which is considered last in the call insertIntoMST($w$). By induction hypothesis, $m$ and $t$ are the largest edges in the paths joining $w$ and $x$ to $z$, respectively, after executing insertIntoMST($x$). It can be shown that in all cases $t$ will be the largest edge in the path joining $w$ to $z$. Similarly, $m$ is the largest edge in the path joining $r$ to $z$.

Also, in lines 6 to 9, the largest edge among $m$, $(w,r)$ and $t$ is deleted, and thus $m$ and $T$ (the MST) are updated. Since at most $n-1$ edges are deleted, each of which was the largest in a cycle, $T$ will still remain a MST.

Because insertIntoMST($r$) has $(n-1)$ recursive calls at line $5$, the lines $1, 2$, $6$--$10$ and $14$ are executed $n$ times. Lines $3$ and $4$ counts each tree edge twice (at most), and as this is proportional to the adjacency list, those are executed at most $2(n-1)$ times. Therefore, Algorithm~\ref{alg:insert_opfi} runs in $\mathcal{O}(n)$.
\end{proof}

\section{Experiments}
\label{s.experiments}

\subsection{Data sets}

The code and all datasets that are not publicly available can be found at \url{http://www.icmc.usp.br/~moacir/paper/16opfi.html}. A variety of synthetic and real datasets are used in the experiments.

    {\bf Synthetic datasets:} Base1/Base2/Base3: with size 10000 and data distributed in 4 regions. Base1 has 2 classes, Base2 has 4 classes and Base3 has 3 classes; Lithuanian, Circle vs Gaussian (C-vs-G), Cone-Torus and Saturn: are 2-d classes with different distributions.
    
    {\bf Real datasets}: CTG cardiotocography dataset, 2126 examples, 21 features, 3 classes; NTL (non-technical losses) energy profile dataset, 4952 examples, 8 features, 2 classes; Parkinsons dataset, 193 examples, 22 features, 2 classes; Produce image dataset, 1400 examples, 64 features, 14 classes; Skin segmentation dataset, 245057 examples, 3 features, 2 classes; SpamBase email dataset, 4601 examples, 56 features, 2 classes; MPEG7-B shape dataset, 70 classes and 20 examples.

\subsection{Experimental setup}

Experiments were conduced to test the OPFI algorithm and comparing with the original OPF and DIFT. We aim to get similar accuracies with respect to them in linear time. Each experiment was conducted in $10$-repeated hold-out sampling:

\begin{enumerate}
\item \textbf{Split data 50-50}: $S$ for supervised training and $T$ for testing, keeping the class distribution of the original dataset;
\item \textbf{Split $S$}: maintaining class proportions: BaseN, C-vs-G, Lithu, CTG, NTL, Parkinsons, Produce, SpamBase, Skin: into 100 subsets $S_i$, with $i=0..99$. Cone-Torus, Saturn, MPEG7: into 10 subsets $S_i$, with $i=0..9$ (fewer subsets because they have few examples/class).
\item \textbf{Initial training on $S_0$} using original OPF as base to be incremented;
\item \textbf{Update model} including each $S_i$ in sequence, starting with $i=1$.
\end{enumerate}

\section{Results and Discussion}

\begin{figure*}
    \begin{center}
        \begin{tabular}{cc} 
            \includegraphics[width=0.38\linewidth]{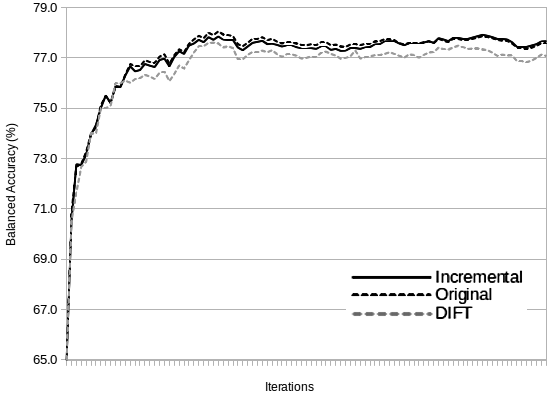} &
            \includegraphics[width=0.38\linewidth]{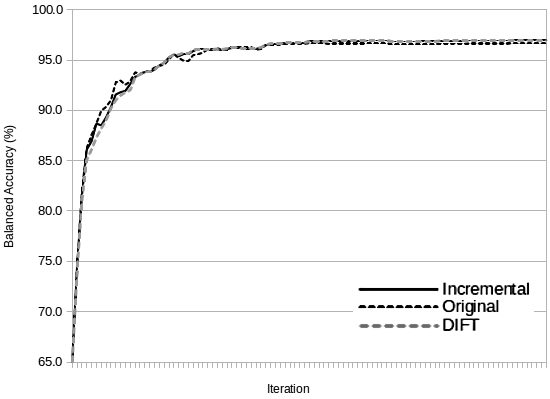} 
            \\
            \includegraphics[width=0.38\linewidth]{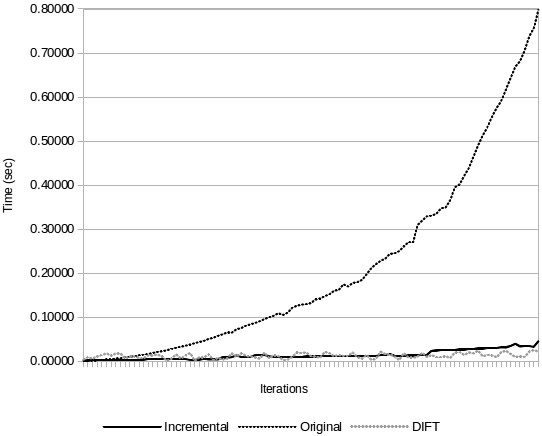} &
            \includegraphics[width=0.38\linewidth]{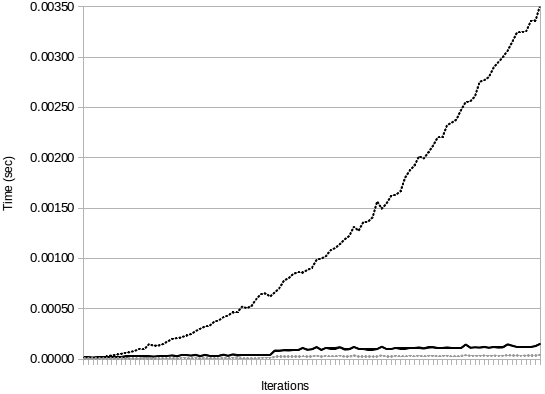} \\[3pt]
            (a) Lithuanian & (b) Circle-vs-Gaussian \\[6pt]

            \includegraphics[width=0.38\linewidth]{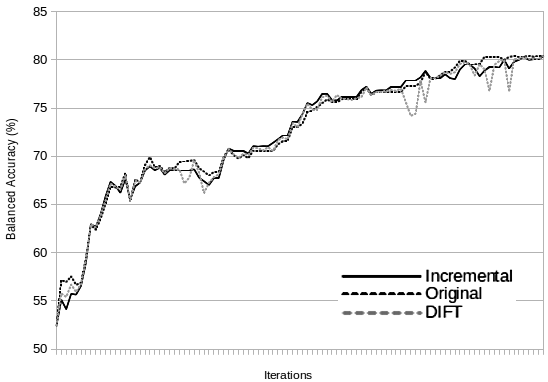} &
            \includegraphics[width=0.38\linewidth]{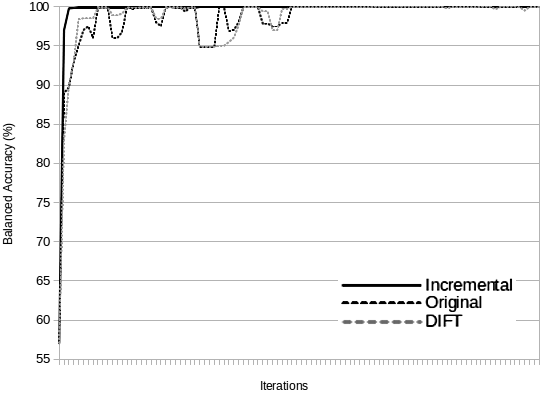} 
            \\
            \includegraphics[width=0.38\linewidth]{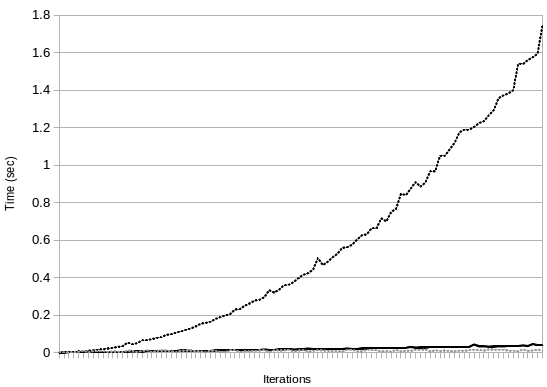} &
            \includegraphics[width=0.38\linewidth]{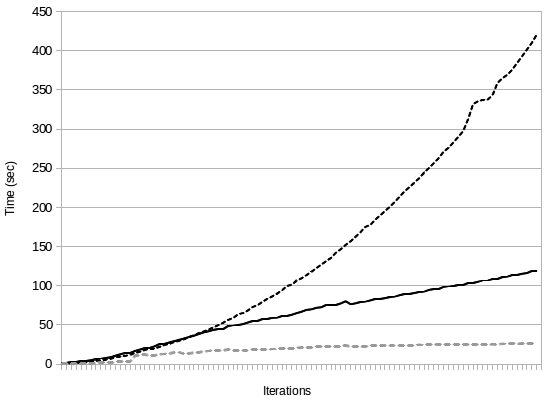} \\[3pt]
			(c) NTL & (d) Skin
        \end{tabular}
        \caption{Balanced accuracies (first and third rows) and running time (second and forth rows) for each iteration on the Lithuanian, Circle-vs-Gaussian, NTL and skin datasets}
        \label{fig:resultados_real}
    \end{center}
    \end{figure*}

The balanced accuracy (takes into account the proportions of examples in each class) results are shown in Table~\ref{tab:dift_real}. A plot with accuracy and running time results for 3 datasets are shown in Figure~\ref{fig:resultados_real}. By inspecting the average and standard deviations, it is possible to see that OPFI is able to keep accuracies similar to original OPF and DIFT. However, it runs faster then the OPF and does not degrade the graph structure as happens with DIFT. The optimum-path trees are preserved and therefore can be explored in scenarios when incremental learning is needed.

The running time curves shows the linear versus quadratic behaviour on all experiments: for $n$ examples in the previous model, each new inclusion with original OPF would take $\mathcal{O}((n+1)^2)$, while with OPFI it is performed in $\mathcal{O}(n+1)$. When including examples in batches, our algorithm runs in $\mathcal{O}(n\cdot b)$, where $b$ is the batch size, while OPF runs in $\mathcal{O}((n+b)^2)$. Therefore our method suits better several small inclusions than large batches. Neverthelles, OPFI's running time $(n\cdot b)$ is still $\mathcal{O}(n)$ and $o((n + b)^2)$.

\renewcommand{\arraystretch}{0.75}    
\setlength\tabcolsep{2pt}

    \begin{table*}
        \caption{Balanced accuracy results for the initial model, the first 3 increments, 50\% and 100\% of the increments}
     {\footnotesize
        \begin{center}
          \begin{tabular}{ccr|rrr|rr} 
                \hline
                        &  & $S_0$       			 &   1st      		 &   2nd  		 &   3rd  		 &  50\%   &   100\%   \\      
                \hline
                \hline
                Base1   & Incremental & $85.3 \pm 4.8$ & $89.6 \pm 1.8$ & $91.8 \pm 0.9$ & $92.9 \pm 0.8$ & $98.0 \pm 0.2$ & $98.5 \pm 0.1$ \\
                        & Original    & $85.3 \pm 4.8$ & $89.3 \pm 2.5$ & $91.1 \pm 1.1$ & $92.3 \pm 0.8$ & $98.0 \pm 0.2$ & $98.3 \pm 0.2$ \\
                        & DIFT    	  & $85.3 \pm 4.8$ & $89.2 \pm 1.8$ & $91.5 \pm 0.8$ & $92.7 \pm 0.7$ & $97.7 \pm 0.3$ & $98.5 \pm 0.1$ \\
                \hline
                Base2   & Incremental & $90.0 \pm 2.2$ & $93.4 \pm 0.8$ & $94.6 \pm 0.6$ & $95.0 \pm 0.6$ & $98.7 \pm 0.5$ & $99.1 \pm 0.1$ \\
                        & Original    & $90.0 \pm 2.2$ & $92.8 \pm 1.0$ & $94.3 \pm 0.9$ & $95.1 \pm 0.9$ & $98.7 \pm 0.7$ & $99.1 \pm 0.1$ \\
                        & DIFT 		  & $90.0 \pm 2.2$ & $93.2 \pm 1.0$ & $94.5 \pm 0.7$ & $94.9 \pm 0.6$ & $98.5 \pm 0.4$ & $99.1 \pm 0.1$ \\
                \hline
                Base3   & Incremental & $88.7 \pm 2.2$ & $92.1 \pm 1.6$ & $93.8 \pm 1.2$ & $94.5 \pm 0.8$ & $98.5 \pm 0.2$ & $98.9 \pm 0.1$ \\ 
                        & Original    & $88.7 \pm 2.2$ & $91.9 \pm 1.6$ & $93.6 \pm 1.1$ & $93.9 \pm 0.8$ & $98.3 \pm 0.2$ & $98.9 \pm 0.1$ \\
                        & DIFT  	  & $88.7 \pm 2.2$ & $91.9 \pm 1.4$ & $93.8 \pm 1.1$ & $94.1 \pm 0.8$ & $98.5 \pm 0.3$ & $98.9 \pm 0.1$ \\
                \hline
                C-vs-G  & Incremental & $64.9 \pm 10.9$ & $74.0 \pm 7.0$ & $81.2 \pm 7.4$ & $85.5 \pm 5.4$ & $96.8 \pm 0.7$ & $97.7 \pm 0.8$ \\
                        & Original    & $64.9 \pm 10.9$ & $74.6 \pm 8.1$ & $81.9 \pm 7.3$ & $86.3 \pm 5.5$ & $96.8 \pm 1.0$ & $96.7 \pm 1.0$ \\
                        & DIFT 		  & $64.9 \pm 10.9$ & $74.0 \pm 8.2$ & $81.0 \pm 6.9$ & $85.0 \pm 4.8$ & $96.9 \pm 0.8$ & $97.0 \pm 0.8$ \\
                \hline
                Lithu   & Incremental & $65.0 \pm 4.9$ & $70.5 \pm 4.4$ & $72.8 \pm 3.0$ & $72.7 \pm 2.9$ & $77.4 \pm 1.2$ & $77.6 \pm 0.6$ \\
                        & Original    & $65.0 \pm 4.9$ & $70.5 \pm 4.0$ & $71.7 \pm 3.1$ & $72.7 \pm 3.4$ & $77.1 \pm 1.3$ & $77.0 \pm 1.0$ \\
                        & DIFT 		  & $65.0 \pm 4.9$ & $70.8 \pm 4.3$ & $72.7 \pm 3.1$ & $72.8 \pm 2.9$ & $77.1 \pm 1.3$ & $76.9 \pm 0.7$ \\
                \hline
                Cone-Torus& Incremental & $81.0 \pm 2.6$ & $83.7 \pm 2.7$ & $85.0 \pm 1.7$ & $85.8 \pm 2.6$ & $87.6 \pm 1.1$ & $87.9 \pm 1.0$ \\
                          & Original    & $81.0 \pm 2.6$ & $82.9 \pm 2.4$ & $84.8 \pm 2.1$ & $85.0 \pm 2.8$ & $87.1 \pm 0.8$ & $87.3 \pm 1.2$ \\
                          & DIFT        & $81.0 \pm 2.6$ & $83.6 \pm 2.6$ & $84.8 \pm 1.6$ & $85.5 \pm 2.7$ & $87.7 \pm 1.1$ & $87.1 \pm 1.3$ \\

                \hline
                    Saturn& Incremental & $58.9 \pm 11.6$ & $68.5 \pm 4.0$ & $73.9 \pm 5.3$ & $78.3 \pm 3.7$ & $81.8 \pm 3.8$ & $88.2 \pm 1.7$ \\
                          & Original    & $58.9 \pm 11.6$ & $69.0 \pm 3.9$ & $74.5 \pm 5.5$ & $78.4 \pm 3.9$ & $82.0 \pm 4.0$ & $88.0 \pm 1.7$ \\
                          & DIFT        & $58.9 \pm 11.6$ & $68.4 \pm 3.2$ & $73.5 \pm 5.4$ & $78.1 \pm 2.4$ & $81.8 \pm 3.7$ & $87.6 \pm 1.3$ \\        
                \hline
                \hline
                CTG     & Incremental   &  $72.2 \pm 8.3$ & $80.5 \pm 2.7$ & $81.0 \pm 3.5$ & $83.0 \pm 2.9$ & $92.8 \pm 0.6$ & $93.9 \pm 0.8$ \\
                        & Original      &  $72.2 \pm 8.3$ & $79.3 \pm 3.0$ & $80.8 \pm 2.9$ & $82.6 \pm 2.7$ & $92.5 \pm 0.7$ & $93.7 \pm 1.2$ \\
                        & DIFT          &  $72.2 \pm 8.3$ & $80.7 \pm 2.9$ & $81.1 \pm 3.5$ & $81.5 \pm 2.9$ & $92.4 \pm 0.6$ & $93.6 \pm 1.1$ \\
                \hline
                NTL     & Incremental   &  $52.4 \pm 3.5$ & $54.1 \pm 1.8$ & $55.5 \pm 2.1$ & $56.9 \pm 3.3$ & $72.2 \pm 0.9$ & $82.0 \pm 0.6$ \\
                        & Original      &  $52.4 \pm 3.5$ & $53.6 \pm 2.0$ & $55.6 \pm 2.0$ & $57.0 \pm 3.3$ & $72.8 \pm 1.0$ & $82.0 \pm 0.7$ \\
                        & DIFT          &  $52.4 \pm 3.5$ & $54.1 \pm 1.8$ & $55.0 \pm 3.0$ & $55.6 \pm 3.3$ & $71.6 \pm 0.9$ & $80.4 \pm 0.7$ \\
                \hline
                Parkinsons & Incremental& $60.9 \pm 11.8$ & $62.0 \pm 10.2$ & $69.7 \pm 12.2$ & $72.6 \pm 7.5$ & $89.0 \pm 3.8$ & $89.4 \pm 3.6$ \\
                           & Original   & $60.9 \pm 11.8$ & $61.4 \pm 12.0$ & $67.5 \pm 11.9$ & $72.1 \pm 8.2$ & $87.1 \pm 5.6$ & $88.1 \pm 5.0$ \\
                           & DIFT 	    & $60.9 \pm 11.8$ & $62.8 \pm 10.2$ & $69.3 \pm 12.3$ & $72.2 \pm 7.7$ & $89.1 \pm 3.7$ & $89.5 \pm 3.4$ \\
                \hline
                Produce  & Incremental  & $63.6 \pm 1.8$ & $70.4 \pm 2.1$ & $74.4 \pm 1.4$ & $77.8 \pm 0.8$ & $95.2 \pm 0.6$ & $95.1 \pm 0.6$ \\
                         & Original     & $63.6 \pm 1.8$ & $70.3 \pm 2.1$ & $74.3 \pm 1.4$ & $77.6 \pm 0.7$ & $95.2 \pm 0.5$ & $95.2 \pm 0.7$ \\
                         & DIFT         & $63.6 \pm 1.8$ & $70.4 \pm 2.1$ & $74.4 \pm 1.4$ & $77.8 \pm 0.8$ & $94.5 \pm 0.6$ & $94.5 \pm 0.6$ \\
                \hline
                SpamBase & Incremental & $71.9 \pm 3.1$ & $76.0 \pm 3.5$ & $78.0 \pm 2.4$ & $78.7 \pm 2.0$ & $85.6 \pm 0.7$ & $87.6 \pm 0.6$ \\
                         & Original    & $71.9 \pm 3.1$ & $75.8 \pm 3.3$ & $77.8 \pm 2.2$ & $78.5 \pm 1.9$ & $85.1 \pm 0.7$ & $87.0 \pm 1.0$ \\
                         & DIFT        & $71.9 \pm 3.1$ & $76.0 \pm 3.6$ & $78.0 \pm 2.4$ & $78.6 \pm 2.1$ & $84.7 \pm 0.7$ & $85.6 \pm 0.3$ \\
                \hline
                MPEG7-B & Incremental & $72.9 \pm 0.8$ & $78.2 \pm 0.9$ & $81.2 \pm 1.0$ & $83.1 \pm 0.9$ & $85.4 \pm 0.6$ & $91.6 \pm 0.4$ \\
                        & Original    & $72.9 \pm 0.8$ & $78.1 \pm 0.9$ & $81.1 \pm 0.9$ & $82.9 \pm 0.8$ & $85.3 \pm 0.6$ & $91.5 \pm 0.4$ \\
                        & DIFT        & $72.9 \pm 0.8$ & $78.2 \pm 0.9$ & $81.2 \pm 1.0$ & $83.0 \pm 0.8$ & $85.5 \pm 0.6$ & $91.7 \pm 0.5$ \\
                 \hline
                 Skin  & Incremental  & $57.0 \pm 1.8$ & $97.0 \pm 1.2$ & $99.7 \pm 0.2$  & $99.8 \pm 0.0$ & $99.9 \pm 0.0$ & $99.9 \pm 0.0$ \\
                       & Original     & $57.0 \pm 1.8$ & $89.0 \pm 1.2$ & $89.7 \pm 0.4$ & $93.5 \pm 0.2$  & $99.8 \pm 0.1$ & $99.9 \pm 0.0$ \\
                       & DIFT         & $57.0 \pm 1.8$ & $83.0 \pm 1.2$ & $90.2 \pm 1.0$ & $93.1 \pm 0.8$  & $99.8 \pm 0.2$ & $99.9 \pm 0.0$ \\

                    \hline
        \end{tabular}
        \label{tab:dift_real}
        \end{center}
        }
    \end{table*}

%
%
We believe our contribution will allow the OPF method to be used more efficiently in future studies, such as data stream mining and active learning applications. Previous algorithms for decreasing the running time of the OPF training step can also be used within each batch of examples to be added to the OPFI algorithm to further speed-up the process.

\section*{Acknowledgment}
We would like to thank FAPESP (\#11/22749-8 and \#11/16411-4).

\bibliographystyle{elsarticle-num}
\bibliography{RivaPontiOPFIncremental}

\end{document}